\documentclass[twoside,11pt]{article}

\usepackage{jmlr2e_mod}
\usepackage{listings}

\usepackage{epsfig}
\usepackage{graphicx}
\usepackage{caption}

\usepackage{mathrsfs}
\usepackage{color}
\usepackage{amssymb}
\usepackage[utf8]{inputenc}
\usepackage{afterpage}
\usepackage[T1]{fontenc}
\usepackage{amsfonts}
\usepackage{mathtools}
\usepackage{amsmath}
\usepackage{amssymb}
\usepackage{verbatim}
\usepackage[margin=1.2in]{geometry}
\usepackage{bm}
\usepackage{bbm}
\usepackage{hyperref}
\usepackage{enumitem}
\usepackage{stmaryrd}
\usepackage{natbib}
\usepackage{lastpage}
\usepackage[boxruled]{algorithm2e}

\newcommand*\mcupinn[2]{\vcenter{\hbox{$\mathsurround=0pt
  \ifx\displaystyle#1\textstyle\else#1\fi\bigcup$}}}

\newcommand*\mcapinn[2]{\vcenter{\hbox{$\mathsurround=0pt
  \ifx\displaystyle#1\textstyle\else#1\fi\bigcap$}}}
\DeclarePairedDelimiter{\abs}{\lvert}{\rvert}
\DeclarePairedDelimiter{\norm}{\lVert}{\rVert}
\DeclarePairedDelimiter{\parr}{(}{)}

\DeclarePairedDelimiter{\bra}{\lbrace}{\rbrace}

\DeclareMathOperator*{\argmin}{arg\,min}
\DeclareMathOperator*{\argmax}{arg\,max}

\DeclareMathOperator*{\st}{\,:\,}

\newcommand{\R}{{\mathbb R}}
\newcommand{\D}{{\mathcal{D}}}
\newcommand{\E}{{\mathcal{E}}}
\newcommand{\X}{{X}}
\newcommand{\hatX}{{\hat{X}}}
\newcommand{\labeled}{{X_\ell}}
\newcommand{\unlabeled}{{X_u}}
\newcommand{\pool}{{X_\mathrm{pool}}}
\newcommand{\p}{{\mathrm{pool}}}

\ShortHeadings{Diffusion-based Deep Active Learning}{Kushnir and Venturi}

\begin{document}

\title{Diffusion-based Deep Active Learning}

\author{\name Dan Kushnir \email dan.kushnir@nokia-bell-labs.com \\
\addr Bell Laboratories, Nokia, Murray Hill, NJ, USA
\AND
\name Luca Venturi \email venturi@cims.nyu.edu \\
\addr Courant Institute of Mathematical Sciences, New York University, New York, NY, USA
}

\maketitle

\begin{abstract}
The remarkable performance of deep neural networks depends on the availability of massive labeled data. To alleviate the load of data annotation, active deep learning aims to select a minimal set of training points to be labelled which yields maximal model accuracy.
Most existing approaches implement either an `exploration'-type selection criterion, which aims at exploring the joint distribution of data and labels, or a `refinement'-type criterion which aims at localizing the detected decision boundaries. We propose a versatile and efficient criterion that automatically switches from exploration to refinement when the distribution has been sufficiently mapped. Our criterion relies on a process of diffusing the existing label information over a graph constructed from the hidden representation of the data set as provided by the neural network. This graph representation captures the intrinsic geometry of the approximated labeling function. The diffusion-based criterion is shown to be advantageous as it outperforms existing criteria for deep active learning.

\end{abstract}

\section{Introduction}
Deep learning has provided unprecedented performance in various semi-supervised learning tasks ranging from speech recognition to computer vision and natural language processing. Deep Convolutional Neural Networks (CNN), in particular, have demonstrated object recognition that exceeds human's performance. However, this success comes with the requirement for massive amounts of labelled data. While data collection has become easier, the annotation of such data with labels is time consuming and expensive. In fact, the annotation of data has become a major bottleneck in the application of deep learning to many real life problems.

Active learning provides a plethora of techniques that allows to select a minimal set of data points for labeling which optimally minimize the error probability under a fixed budget of a labeling effort (see \citep{settles2012active} for a review). A well known trade-off in active learning is between the exploration and refinement stages. Exploration aims at mapping the joint data and labeling distribution in order to identify decision boundaries, and typically yields optimal results at the earlier stage of active learning. Refinement (also referred to as exploitation), on the other hand, characterizes acquisition of labels at the proximity of a discovered decision boundary. Refinement typically provides better gains in accuracy when it follows an exploration stage. It has been shown that active learning exhibits improved results when the balance between exploration and refinement is optimal (e.g. \citep{explVSexplt}).

Incorporating active learning into deep learning is yet a challenging task for several reasons. First, the network representation does not allow to construct an easy probabilistic criterion that incorporates exploration and refinement. Thus, most existing works in active deep learning focus on either exploration (e.g. \citep{sener2017active}) or on refinement type criteria (e.g. \citep{gal2017deep}), which often leads to sub-optimal results. Second, training of deep networks is a costly operation. Hence, acquiring only a small amount of labels at each step is impractical in terms of the overall  running time of retraining the model sequentially. Therefore, at each step a sufficiently large batch of labels needs to be queried. This batch of points needs to be diverse enough to provide higher gains in accuracy for the labeling effort spent. Selecting such a batch may be a daunting task, as often a model needs to be trained to generate prediction for the unlabeled set  while a candidate set of labels is hypothesized, as, for example, in the model-change criterion for active learning \citep{settles2012active}, or in \citep{SettlesEMC}. This state pushed active deep learning to focus mostly on simple uncertainty-based criteria (e.g. \citep{gal2017deep,wang2016cost,stark2015captcha}), or geometric criterion (e.g. \citep{sener2017active}) to avoid at least some of the re-training burden. 

In this paper we propose a batch active learning criterion that uses a label diffusion process on a graph constructed from the hidden layer representation. The main contributions of this approach are with addressing the solution of the above problems. First, the label diffusion based query criterion exhibits a natural switch from exploration to refinement, which avoids the need for an additional optimization machinery (e.g. as in \citep{Yang}). It has the versatility of an explorer and a refiner by using just a single query criterion. Moreover, the learning of the data distribution is done on a graph constructed from the top hidden layers which are highly correlated with the labeling approximation. The graph representation captures the intrinsic geometry of the labeling function instead of probing it in the ambient feature space as, for example, in \citep{sener2017active}.

To address the second problem above, we note that label diffusion has running time that is linear in data size as it involves the application of a sparse matrix to a sparse vector for only a handful of iterations. The graph construction is based on a nearest neighbor procedure, which can be efficiently done if the hidden layer representation is low dimensional, as typical to the top layers. Once the diffusion iterations are done, the subsequent query selection step involves nothing but a quick sort of the diffused values. Therefore, the batch querying process is {\color{blue}  } fast, can scale to {\color{blue}  } and diverse batches, and does not require model training, as for example in model-change-based criteria.

The optimality of our criterion is demonstrated on benchmark data over a range of state-of-the-art active learning criteria and in particular those constructed for deep learning.

\section{Related work}

Active learning is a well-studied branch of machine learning. To reflect on the diverse approaches used in active learning we note some of the earlier seminal works of \citep{gale} on using posterior probability for uncertainty sampling, \citep{MacKay} using information theory, \citep{Tong} for using geometric distance to decision boundary, \citep{qbc,Dagan} which suggested an ensemble committee of classifiers, and \citep{McCallum} using expected risk estimation. For an introduction, we refer to the comprehensive review \citep{settles2012active}.

In light of the important impact of active learning, it has recently been incorporated with deep models. In this context, a first class of active learning criteria are \emph{uncertainty} based method. Such methods sample points from the unlabeled data set according to some measure of uncertainty, extracted from the so-far learned classifier. A few recent works \citep{stark2015captcha,wang2016cost, Wang} considered these methods to learn deep neural nets. Recently, the authors of \citep{ducoffe2018adversarial} proposed \emph{margin}-based approach for deep learning. In \citep{ducoffe2018adversarial}, the distance of data points from the decision boundaries is estimated as the distance to the closest adversarial examples. Points closer to the decision boundaries are queried. While uncertainty methods can be effective at later stages of active learning, when the learned model is already a discrete classifier, their performances can be even lower than random labeling at earlier stages, as we show in Section \ref{section:experiments}.

Another important family of active learning techniques is the one of \emph{ensemble} methods. In this setting the uncertainty measure is computed over an ensemble of models trained on the existing labeled dataset \citep{Beluch}. Due to the high computational effort of training deep neural networks, such methods can be impractical. A recent work-around has been recently proposed in \citep{gal2017deep}, based on an equivalence between dropout and approximate Bayesian inference \citep{gal2016dropout}. In this work the authors average the uncertainties over the outputs of an ensemble of networks obtained via dropout.

\emph{Exploratory} criteria have also been experimented for learning deep neural networks. The works \citep{sener2017active} and \citep{geifman2017deep} proposed to perform large batch queries by choosing data that form an $\epsilon$-covering of the data distribution. These methods were shown to yield gains in some cases, but they seem to be less effective for querying smaller batches at a time and they typically require an expensive optimization problem.
In a recent work a discriminative network was trained in \citep{gissin2019discriminative} as a proxy for exploration. 
While promising, these methods seemed to perform similarly on some benchmark data sets, as discussed in \citep{gissin2019discriminative}. 

We believe that a key concept missing in the above cited methods is the combination of exploration and refinement criteria in active learning. The above cited works either try to tackle the decision boundaries based on the current model or to cover the data distribution. In the deep learning setting, a first attempt to merge the two types of criteria was proposed in \citep{liu2017active} where the authors use a heuristic method to combine uncertainty with representativeness designated for the specific task of human pose annotation. We introduce in this paper a versatile and principled active learning criterion based on label-diffusion over graphs. Label-diffusion based criterion was first proposed in \citep{kushnir2014active}, where it was coupled with a label-adaptation of the graphs diffusion weights to mitigate diffusion in the label space. Here we do not require the label adaptation of the diffusion kernel, as the graph is constructed from the top hidden layer, which is highly correlated with the hypothesized labels. 


\section{Problem set-up} \label{sec:setup}

We consider the following multi-class classification set-up. Let $\D$ be a generic probability distribution over $\R^d\times [C]$, where $C\geq 2$ indicates the number of classes. Our aim is to find a classifier $f:\R^d \to \Delta_C$ ($\Delta_C$ denotes the space of probability measures over $[C]$) that minimizes the classification error
$$
\mathcal{E}(f) = \mathbb{P}_{(x,y) \sim \mathcal{D}}\bra*{ \argmax_{c} f_c(x_i)\neq y_i }
$$
In deep learning, we consider the parametric functions
$$
f_\theta = g_{\theta_{n+1}} \circ h^n_{\theta_{n}} \circ \cdots \circ h^1_{\theta_{1}}
$$
where $h^i_{\theta_i} : \R^{d_{i-1}} \to \R^{d_{i}}$ (with $d_0 = d$) and $g_{\theta_{n+1}}:\R^{d_n}\to \Delta_C$. The parameter $\theta = \bra{\theta_i}_i$ defines our final model $f_\theta$. In particular, in the following experiments, $f_\theta$ is taken to be a neural network. In this case the function $h^i_{\theta_i}$ corresponds to the $i$-th layer of the network and $g_{\theta_{n+1}}$ to the final classifier layer.

\paragraph{Active learning}

In active learning, one is given a \emph{pool} of input data points $\pool = \bra{x_i}_{i=1}^N$, and a budget $Q$ of data points to select for labeling. Optionally, for a subset $\labeled \subset \pool$, the labels of the data points may also be given as input. In the following, we denote $\unlabeled \doteq \pool \setminus \labeled$ as the set of unlabeled points in $\pool$.
The learner is given the possibility to query from an oracle the labels of up to $Q$ points in $\pool$. At each iteration of active learning, the newly labeled points are added to the existing set of labeled points, and the model can be retrained with the updated training set. The aim of active learning is to minimize the approximation error $\E(f_\theta)$ while querying at most $Q$ points from $\pool$.
An active learning strategy consists of the following steps:
\begin{enumerate}
\item Train $f_\theta$ on a given set of labeled samples $\X_\ell$.

\item Until a given budget $Q$ of labels is queried, repeat:
\begin{enumerate}
\item Select a subset of data points $\hatX \subset \X_u $ using a certain \emph{selection criterion}.
\item Query the labels of the data points in $\hatX$ from an oracle and add them (with their labels) to $\X_\ell$: $\X_\ell = \hatX  \cup \X_\ell$.
\item Train the model $f_\theta$ on the augmented set of labeled points $\X_\ell$.
\end{enumerate}
\end{enumerate}

The task in this paper is to design an efficient and meaningful criterion for data labeling in the above setting.

\section{Classification via Graph Diffusion}
\label{section:graph_construction}
In this section, we describe a framework for classification based on label diffusion over graphs. This framework is at the core of the diffusion-based active learning criterion that we introduce for deep learning in Section \ref{section:algorithm}.

We assume a finite weighted graph $G =(V,E)$ consisting of a set of $N$ \emph{vertices} $V$, a set of \emph{edges} $E\subset V\times V$, and a non-negative \emph{weight} function $W : E \rightarrow [0,\infty)$. 
We interpret the weight $W_{ij} = W(v_i, v_j)$ as a measure of similarity between the vertices $v_i$ and $v_j$, for  $(i,j) = (v_i,v_j)\in E$.
The graph {kernel} is defined by
\begin{equation}
M\doteq D^{-1}W
\label{eq:kernel}
\end{equation}
where $D$ is diagonal with $D_{ii}= \sum_{j}W_{ij}$. The operator $M$ is a weighted averaging operator over functions $f$ defined over the graph: $f_i=D_{ii}^{-1}\sum_{(i,j) \in E} W_{ij}f_j$, with the weights given by the similarities $W_{ij}$. It has the `averaging effect' of \textit{smoothing} the function $f$ over the graph.

In the data context, a graph $G$ can be constructed in which the vertices of $G$ correspond to the data points in $\pool$. The weights $W$ represents similarities between data points:
\begin{equation*}
W_{ij} = m\left(\frac{\rho(x_i,x_j)}{\sigma_{ij}}\right),
\end{equation*}
where $\sigma_{ij}$ is a local scaling parameter, $m$ is a decreasing function, and $\rho$ is a measure of distance between data points. The actual similarities used are the local ones in order to preserve local geometry and reduce computation time by using sparse matrices. These local similarities are realized by computing the \textit{K}-nearest-neighbors for each point $x$, denoted by $N(x)$.

\subsection{Label diffusion on graphs}\label{section:diffusion}
At the center of our algorithmic ideas lies a Markov process that is used to propagate labels from $\labeled$ to $X_u$. 
The graph kernel defined in \eqref{eq:kernel} is a row-stochastic matrix, which can be viewed as the transition probabilities of a Markov random walk on the graph $G$. Specifically, the one-step transition probability between states $x_i$ and $x_k$ is given by $p_{ij} = \mathbb{P}\bra*{x_i \to x_j} = M_{ij}$.



We consider a random walk as a mean to assign a label to $x_i \in X_u$. The predicted label of $x_i$ is associated with the probability of arriving to a labeled point $x$ of class $1$ after performing a $t$-step random walk starting at $x_i$ \citep{Lafferty2} (we consider here the binary case). Marking this probability as $p_t(y(x)=1|i)$, it can be derived by the recursive relation
\begin{equation}
p_t(y(x)=1|i)=\sum_{j}p_{t-1}(y(x)=1|j)p_{ij}.
\label{eq:rwalk}
\end{equation}
We associate $p_t(y(x)=1|i)$ with the probability $p(y(x_i)=1|x_i)$. For labeled points in $\labeled$, $p(y(x_i)=y_i|i)=1$. Denoting $2p_t(y(x)=1|i)-1$ by $\chi_i$ we see that $\chi_i\in [-1,1]$ and its sign can be used to generate binary labels. The vector $\chi$ can be partitioned as $\chi=[\chi_\ell,\chi_u]$, where $\ell$ corresponds to the indices of the labeled nodes in $\labeled$ and $u$ corresponds to the indices of the unlabeled nodes $\pool\setminus \labeled$. Similarly, $D$ and $W$ can be partitioned into blocks
$$
D=\left( \begin{array}{cc}
D_{\ell\ell} & 0 \\
0 & D_{uu} \end{array} \right)\,,\quad   W=\left( \begin{array}{cc}
W_{\ell\ell} & W_{\ell u} \\
W_{u\ell} & W_{uu} \end{array} \right)
$$
Eq. \eqref{eq:rwalk} can be transformed and re-written for $\chi_u$ as
\begin{equation*}
\chi_u= [D_{uu}^{-1}W_{u\ell}\,|\, D_{uu}^{-1}W_{uu}] \parr*{
\begin{array}{c}
\chi_\ell\\
\chi_u\\
\end{array}
}
\end{equation*}
resulting in the system
\begin{equation}
L_{uu}\chi_u=W_{u\ell}\chi_\ell,
\label{eq:lap_system1}
\end{equation}
where $L=D-W$ is the graph Laplacian, and the sign of each $\chi_i$ provides the predicted hard label of $x_i$. A similar system, motivated by quadratic energy minimization, is obtained by minimizing
\begin{equation}
C(\chi)=\frac{1}{2}\sum_{i,j=1}^{l+u}W_{ij}({\chi}_i-{\chi}_j)^2={\chi}^TL{\chi}
\label{eq:energy_func}
\end{equation}
while forcing equality on the labeled set $\chi_\ell = Y_\ell$  \citep{Chapelle}. Specifically, minimizing (\ref{eq:energy_func}) with respect to ${\chi}_u$ leads to
\begin{equation}
L_{u\ell}Y_l+L_{uu}{\chi}_u=0 \quad\Leftrightarrow\quad L_{uu}{\chi}_u=-L_{u\ell}Y_\ell
\label{eq:lap_system2}
\end{equation}
which is the same as (\ref{eq:lap_system1}), since $L_{ul}=-W_{ul}$.

The system (\ref{eq:lap_system2}) can be solved via the well-known Jacobi method \citep{Chapelle}. The iterative Jacobi method solves the system $Az=b$ by approximating the solution at the step $t+1$ by
\begin{equation}
z^{(t+1)}_i=\frac{1}{A_{ii}}\parr*{ b_i-\sum_{j\neq i}A_{ij}z_j^{(t)} }.
\label{eq:jacobi}
\end{equation}
The Jacobi iteration matrix is defined as $B_J=D^{-1}(R+Q)$, where $D$ is a diagonal matrix with $A_{ii}$ on its $i$-th diagonal element, and $R$ and $Q$ are the upper and lower triangular matrices of $A$. In matrix notation the iteration scheme is
\begin{align*}
z^{(t+1)} & = D^{-1}(b-(R+L)z^{(t)})  = B_J z^{(t)} +D^{-1}b .
\end{align*}
For the system (\ref{eq:lap_system2}) we have $A=L_{uu}$, $z=\chi_u$, and $b=L_{u\ell}Y_\ell$, which then yields the iteration
\begin{equation}
{\chi}^{(t+1)}_i=\frac{1}{L_{uu,ij}}\left(-(L_{u\ell}Y_\ell)_i-\sum_{j\neq i}L_{uu, ij}\,{\chi}_j^{(t)}\right).
\label{eq:jacobi_iter}
\end{equation}
Since training points labels are restarted to their true values after every iteration an equivalent system can be considered with $A=L$, $z=[\chi_\ell,\chi_u]$, and $b=[Y_\ell, W_{u\ell}Y_\ell]$ which yields the same result. It is clear now that (\ref{eq:jacobi_iter}) is a label diffusion process: transducing a label to $x_i$ as a weighted average of the labels of its neighbors with the transition weights. 

\subsection{Convergence}
The equivalence we have just drawn between our label propagation and the Jacobi iteration plays an important role in deriving convergence and solution properties for the diffusion process that we propose for active learning.

\begin{lemma}
The iteration in  \eqref{eq:jacobi_iter}
\begin{equation*}
{\chi}^{(t+1)}_i=\frac{1}{L_{uu,ij}}\left(-(L_{u\ell}Y_\ell)_i-\sum_{j\neq i}L_{uu, ij}\,{\chi}_j^{(t)}\right)
\end{equation*}
converges as $t\rightarrow \infty$.
\end{lemma}

\subsection{Algorithm}
Let $\X_\ell^0$ be the available labeled subset of $\pool$ and $f_{\theta^{(0)}}$ a model which has been trained on $\X_\ell^0$. Our goal is to select a batch of unlabeled points $\hatX \subset \X_u^0$ of a given size $\abs{\hatX} = B$ to be labeled and added to the training set such that the accuracy of a model retrained with $\X_\ell^1 = \X_\ell^0 \cup \hatX$  is maximized on $\D$. To facilitate our presentation we first describe in detail the basic query step of our algorithm.  On the following section \ref{sec:enhncd_batch} we address the mechanism to select a large and diverse batch size. 

\paragraph{}

We start by constructing a weighted graph $G = (V,E,W)$ as outlined in Section \ref{section:graph_construction}, with
$$
\hat{W}_{ij} = \exp\bra*{- \frac{\rho(x_i,x_j)}{\sigma_{ij}} }\mathbbm{1}\bra*{j \in N(i)},
$$
where 
\begin{align*}
\rho(x_i,x_j) & = \norm{f^n_{\theta^{(0)}}(x_i) - f^n_{\theta^{(0)}}(x_j)}_2^2, \\ \sigma_{ij} & = \max_{j \in N(i)} \rho(x_i,x_j).
\end{align*}
The neighborhoods $N(i)$ are determined as $K$-NN neighborhoods based on the distance $\rho(x_i,x_j)$ above, where $f_{\theta^{(0)}}^n$ represents the penultimate layer output of the network $f_{\theta^{(0)}}$. Therefore, the constructed graph corresponds to a weighted $K$-NN graph over the set of represented data points $\bra{f_{\theta^{(0)}}^n(x_i)}_{i=1}^N$. 

\paragraph{}

The next step consists of propagating the uncertainty over the graph as suggested in Section \ref{section:diffusion}. The diffusion iteration starts with setting the values in the vector $\chi^{(0)}_c\in\R^N$ (one for each class $c\in[C]$) as
$$
\chi^{(0)}_{i,c} = \begin{cases}
1 & \text{if } i \in \X_\ell \text{ and } c = y_i \\
-1 & \text{if } i \in \X_\ell \text{ and } c \neq y_i \\
0 & \text{if } i \in \X_u.
\end{cases} 
$$
The values $\chi^{(0)}_c\in\R^N$ are propagated as
$$
\chi^{(t+1)}_{i,c} = 
\begin{cases}
1 & \text{if } i \in \X_\ell \text{ and } c = y_i \\
-1 & \text{if } i \in \X_\ell \text{ and } c \neq y_i \\
\parr*{M\chi^{(t)}_{:,c}}_i & \text{if } i \in \X_u
\end{cases}
$$
for $t = T$ time steps. In our supplementary material we propose possible variants on the initialization of $\chi^{(0)}_c$.   

\subsection{Setting $K$ and $T$.}
\label{sec:setTK}
The parameters $T$ and $K$ determine the level of confidence imposed by the diffused training set over the unlabeled set. Higher $K$ impose strong confidence in the labeling hypothesis over larger neighbourhood around each labeled point, but makes the method less computationally efficient (as it controls the sparsity of $M$). 
Similarly, large number of iterations $T$ enables communication between further nodes, but $T$ too large may result in an overly smoothed (and thus less informative) signal $\chi^{(T)}$. During the exploration stage large $T$ imposes an hypothesis that may be locally correct but is still far from being globally reliable, as too few labeled samples cannot correctly approximate the labeling function.

Denote the average number of unlabeled points per a labeled point as $\Gamma =\frac{|X_u |}{|X_l|}$. Since in active learning settings we assume that the labeled set is minimal we have that $\Gamma=O(N)$. For some degree $K$ we can approximate $\Gamma\approx K^T$, with the intuition that the labeling assignment can be recovered by diffusion from the labelled set $X_l$ to all unlabeled nodes $X_u$ in $T$ iterations. Hence, $T=\frac{\log(\Gamma)}{\log(K)}=O(\frac{\log(N)}{\log(K)})$.

The interplay between $K$ and $N$: although $K$ can be treated as a constant, it is preferred that the graph will be connected, otherwise, the exploration phase of the algorithm will be excessively used to label each disconnected component. Hence, if we assume a characteristic graph model (such as Erd\H{o}sh-R\'{e}nye), with degree $k=2\log(N)$ the graph will be connected with high probability.
In our experiments, we use $T \simeq \log_K N$, and $K$ large enough to allow the graph to be connected, as per the above. 
We further discuss on the choices of $K$ and $T$ in the supplemental material.


We conclude that label diffusion is a convergent iteration which provide the label probabilities for each node.

\section{Active Learning}\label{section:algorithm}
In this section we provide the full description of the diffusion-based active deep learning algorithm. The algorithms is based on the graph diffusion framework presented in Section \ref{section:diffusion}, and extended to the batch querying in the deep network setting.

\noindent \paragraph{Query criterion.}
To this end the matrix $\chi^{(T)}$ of propagated values can be interpreted as \emph{uncertainties} measured by the absolute value $\abs{\chi^{(T)}_{c,i}}$. Specifically, the absolute value magnitude represents a measure of uncertainty on whether vertex $i$ belongs to class $c$.  
This can be used to select the new batch to query as
\begin{equation}\label{eq:query}
\hatX = \argmin\nolimits^B_{i \in \X_u} \min_{c\in[C]}\abs*{\chi^{(T)}_{c,i}}
\end{equation}
Here $\min^B$ denotes the $B$ smallest elements. 



The main idea behind the criterion in (\ref{eq:query}) is that it automatically switches between  exploration and refinement stages when exploration provides only little accuracy gains.
In the first exploratory phase, very few labeled points reside sparsely on the graph. Following the diffusion process, the unlabeled points with small label magnitude correspond to unexplored regions that reside far from the labeled set on the data manifold. By querying such data points we ensure to cover the data distribution efficiently.
At later stages, when the data distribution has been explored, the criterion becomes a refiner. At this point, the data points in close proximity to decision boundaries tend to receive the same amount of signal from points of different classes and present smaller label magnitude. Queries that focus on such data points refine the existing decision boundaries.

\paragraph{Running time}

The running time of a batch query is composed of three parts. 
First, we need to compute the $K$-NN graph. This is the most expensive operations and brute force algorithms come with a complexity of $O(dN^2)$, where $d$ is the dimensionality of the data used to compute distances. Using the penultimate layer of the network to represent the data allows us to notably reduce this computational cost. Indeed, in this way we can use procedures for $k$-NN based on KD \citep{bentley1975multidimensional} or ball trees \citep{omohundro1989five}, which come with a computational cost of approximately $O(dN\log_2 N)$. Moreover, at larger stages of active learning, the network representation is closer to be linearly separable, and this also notably speeds up this construction. In practice, in our experiments, we have been using the $K$-NN graph implementation offered in the  Scikit-learn library \citep{scikit-learn}.
Other alternatives include approximate search (e.g. \citep{datar2004locality}) which suggests a trade-off of an almost linear time complexity with a prescribed error constant $\epsilon$.

Second, the diffusion vector is diffused with the kernel matrix. Addressing its sparsity as $O(KN)$ non-zero entries, this operation scales linearly in $N$ as $O(TKN).$ Using our arguments in section \ref{sec:setTK}: $T\simeq O(\log_K N)$, which leads to an $O(KN\log_KN)$ operations for the diffusion stage. 

Finally, the $B$ smallest soft-labels need to be queried. This requires a quick-sort to be applied, which can be done in $O(N\log_2 N)$. We conclude that the running time is $O(dN\log_2N + KN\log_KN + N\log_2N)$.

\subsection{Enhanced large batch queries}
\label{sec:enhncd_batch}
When applying deep active learning to massive data sets, querying large batch sizes is more of a necessity than a choice. For example, querying large size batches using \eqref{eq:query} could possibly results in over-sampling certain regions of $\D$, and wasting labeling efforts. In order to avoid over sampling, we propose to split the batch query in a series of $R$ mini-batch queries of much smaller size $P$ (with $R= B/P$). After each mini-query,  the diffusion vector is updated with the newly added mini-batch in order to enhance the diversification of the next mini-queries. 

The batch query procedure \textit{BatchQuery} for large batches is summarized in Algorithm \ref{algorithm:diffusion_AL_enhanced}. \textit{BatchQuery} takes as input the set of labeled points $\X_\ell^0$, the model $ f_{\theta^{(0)}}$, the batch size $B$, and the mini-batch size $P$. At start, the $K$-NN graph is constructed as outlined in Section \ref{section:algorithm}. Then, for $i = 1,\dots,R=\frac{B}{P}$, we diffuse the labels in $\X_\ell^0 \cup \hatX^{i-1}_\ell$ and use the obtained vector $\chi^{(T)}$ to perform the mini-batch query:
\begin{align}
\mathcal{S} & = \argmin\nolimits^P_{k \in \hatX_u^i } \min_{c\in[C]}\abs*{\chi^{(T)}_{k,c}},
\end{align}
to update $\hatX^{i}_\ell = \mathcal{S} \cup \hatX^{i-1}_\ell$, 
where $\hatX_u^i \doteq \X_\p \setminus\parr*{\X^0_\ell \cup \hatX^{i-1}_\ell}$ is the set of unlabeled data points at iteration $i$. After each mini-batch iteration the labeled set is updated with the new mini-batch, until $R$ mini-batches are queried. We note that the graph $G$ stays unchanged during this process.

 The batch query criterion is summarized in Algorithm \ref{algorithm:diffusion_AL_enhanced}. The full active learning procedure is summarized in Algorithm \ref{algorithm:AL}.

\begin{algorithm}[htb]
\SetKwFunction{Fq}{BatchQuery}
\SetKwProg{Fn}{def}{:}{\KwRet $\hatX^R_\ell$}
\Fn{\textbf{BatchQuery}{$(\pool, \X^0_\ell, f_{\theta^{(0)}}, B, P, K,T)$}}{
    \textbf{input}: $\pool = \bra{x_i}_{i=1}^N$, labeled subset $\X_\ell^0$, model $f_{\theta^{(0)}}$ trained on $\X_\ell^0, \; B,\; P,\;K,\;T$ \\
    \textbf{compute} $K$-NN graph $G = (V,E,W)$ using the top hidden layer representations $\bra{f_{\theta^{(0)}}^n(x_i)}_{i=1}^N$ \\
    \textbf{initialize} $\hatX^{0}_\ell = \emptyset$ \\
    \For{$i = 1:R$}{
    \textbf{initialize} $\chi^{(0)}$ according to $\X_\ell^0 \cup \hatX_\ell^{i-1}$ \\
    \For{$t = 1:T$}{
    $\chi^{(t)}  =  M \chi^{(t-1)} 
    $ \\
    $
    \chi^{(t)}  =  \chi^{(t-1)}|_{\parr*{\X_\ell^0 \cup \hatX_\ell^{i-1}}}
    $
    }
    \textbf{query} $\mathcal{S}  = \argmin\nolimits^P_{k \in \hatX_u^i } \min_{c\in[C]}\abs*{\chi^{(T)}_{c,k}}$ \\
    \textbf{update} $\hatX^{i}_\ell = \mathcal{S} \cup \hatX^{i-1}_\ell$ \\
    }
}
\caption{Batch Query criterion\label{algorithm:diffusion_AL_enhanced}}
\end{algorithm}

\begin{algorithm}[htb]

\textbf{initalization}: $\pool = \bra{x_i}_{i=1}^N$, labeled subset $\X_\ell$, model $f_\theta$ trained on $\X_\ell$, $B,\;P,
\;K\;T$, budget $Q$ \\
\While{$Q>0$}{
\textbf{query} $\hatX = \bm{BatchQuery}(\pool, \X_\ell, f_{\theta}, B,P,K,T)$ \\
\textbf{update} $\X_\ell = \X_\ell \cup \hatX$ \\
\textbf{train} model $f_{\theta}$ on augmented $\X_\ell$\\
\textbf{update} $Q = Q -B$
} 
\caption{Diffusion-based deep active learning \label{algorithm:AL}}
\end{algorithm}

\section{Experiments}
\label{section:experiments}
Our experiments include a demonstration on the trade off between exploration and refinement with a 2-dimensional toy data set, showing the advantages of our versatile query criterion over other  approaches.  We then follow with a set of experiments on the benchmark data sets MNIST, CIFAR10 and SVHN, where we compare other query criteria and algorithms to our diffusion-based
criterion. We conclude with a discussion on the experimental results.

\subsection{Exploration vs. refinement: a toy example}
\begin{figure}[tbp]
\centering
\begin{minipage}[c]{.23\textwidth}
\includegraphics[width=\textwidth,
keepaspectratio]{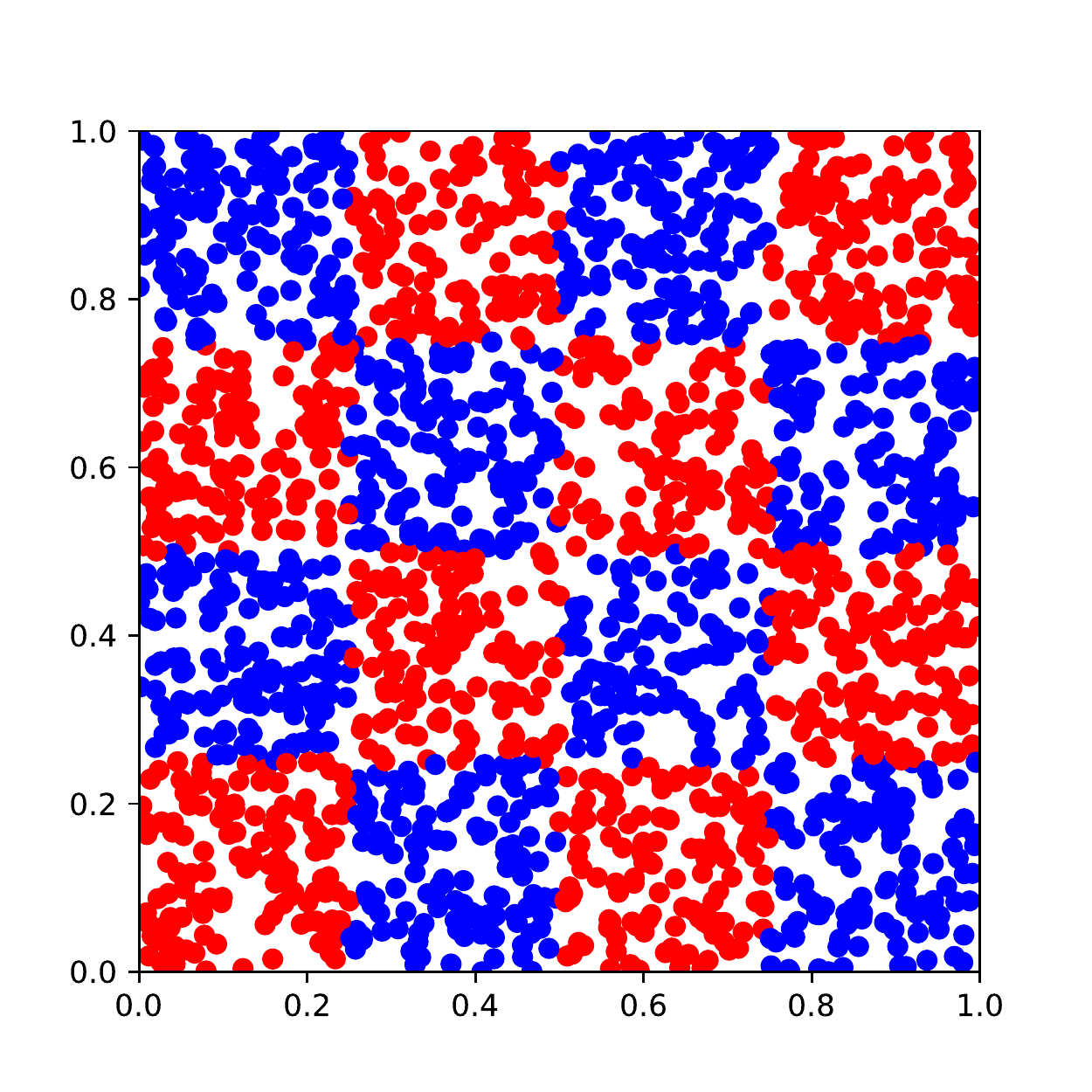}\caption*{Binary Checkerboard}
\end{minipage}
\begin{minipage}[c]{.23\textwidth}
\includegraphics[width=\textwidth,
keepaspectratio]{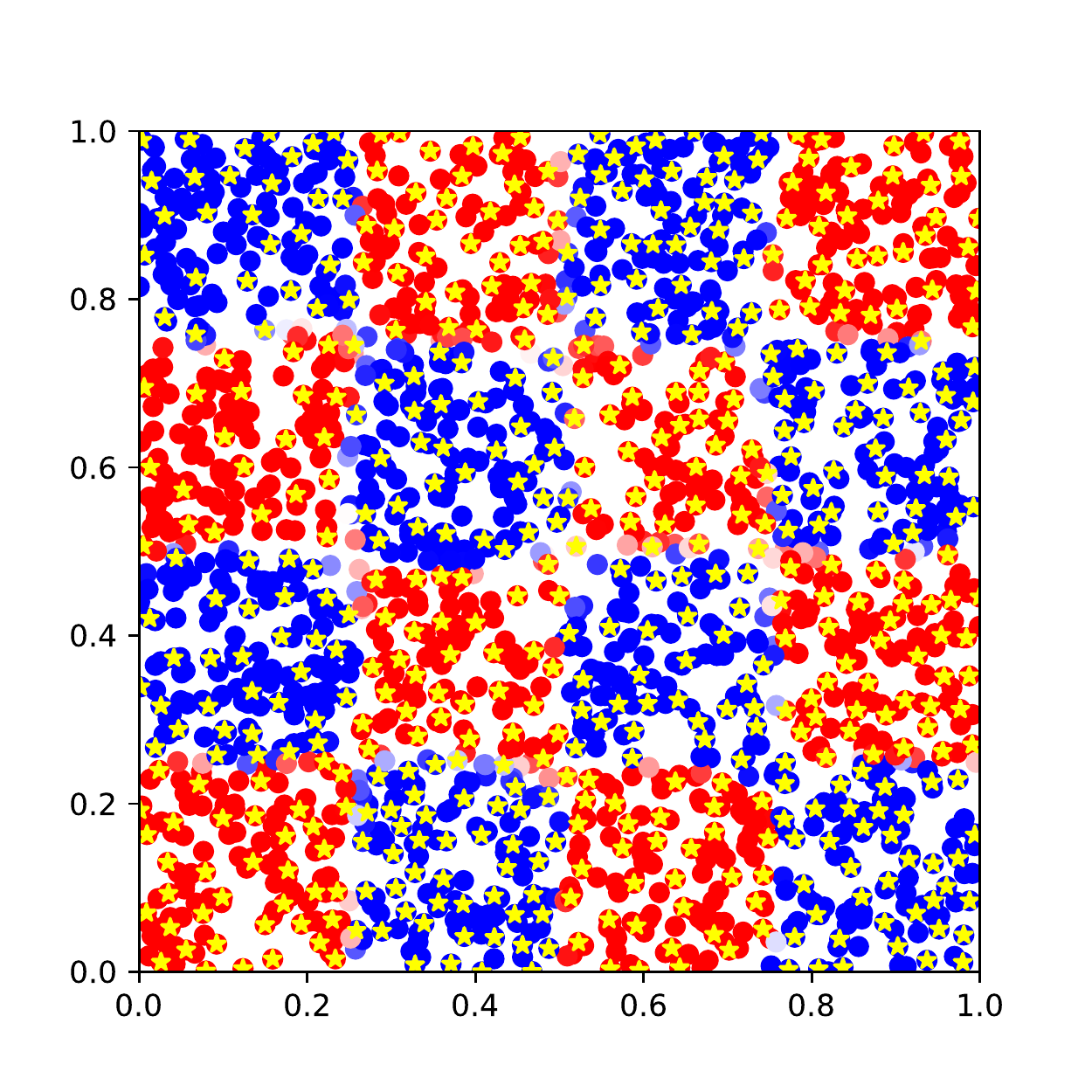}\caption*{Coreset}
\end{minipage}
\begin{minipage}[c]{.23\textwidth}
\includegraphics[width=\textwidth,
keepaspectratio]{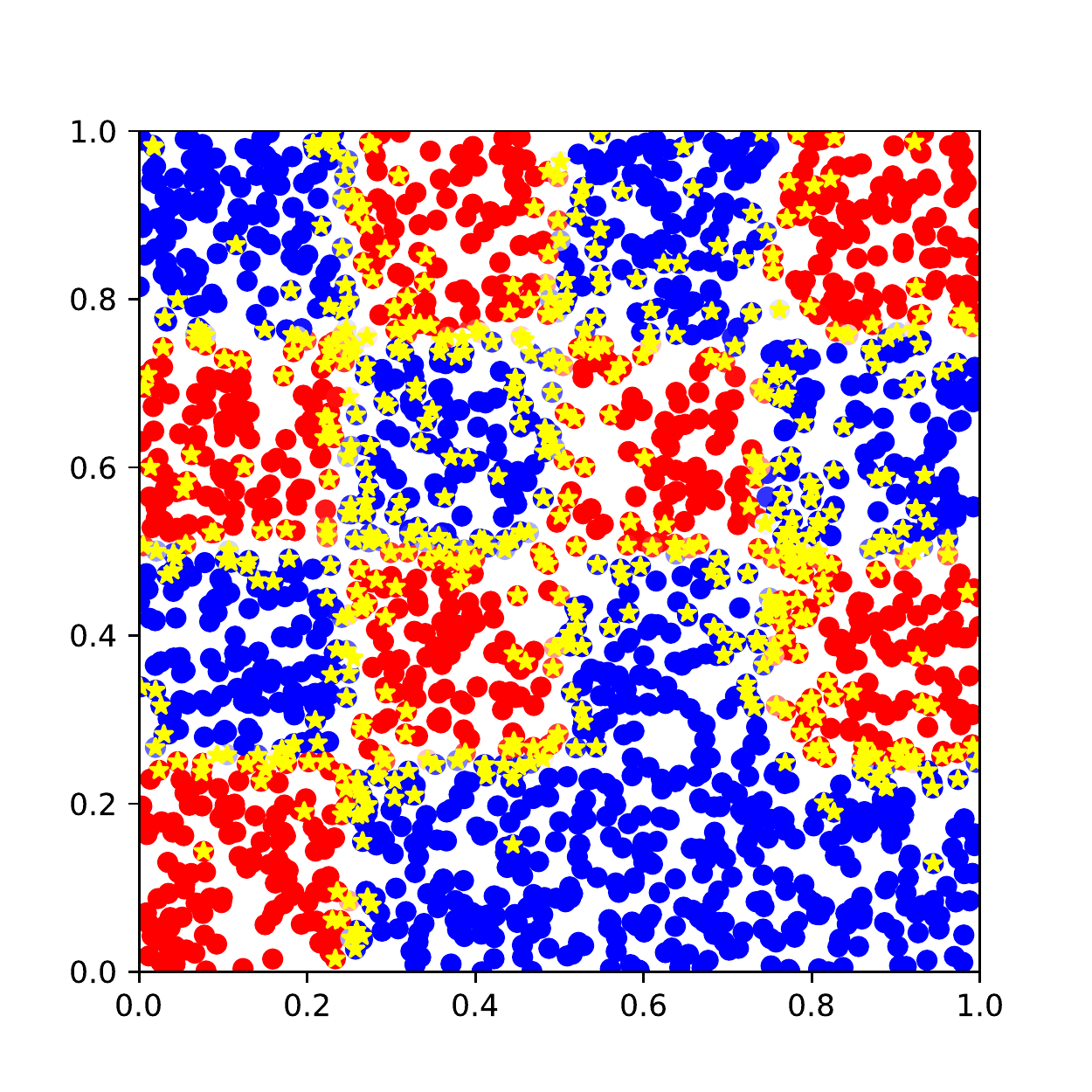}\caption*{Uncertainty}
\end{minipage}
\begin{minipage}[c]{.23\textwidth}
\includegraphics[width=\textwidth,
keepaspectratio]{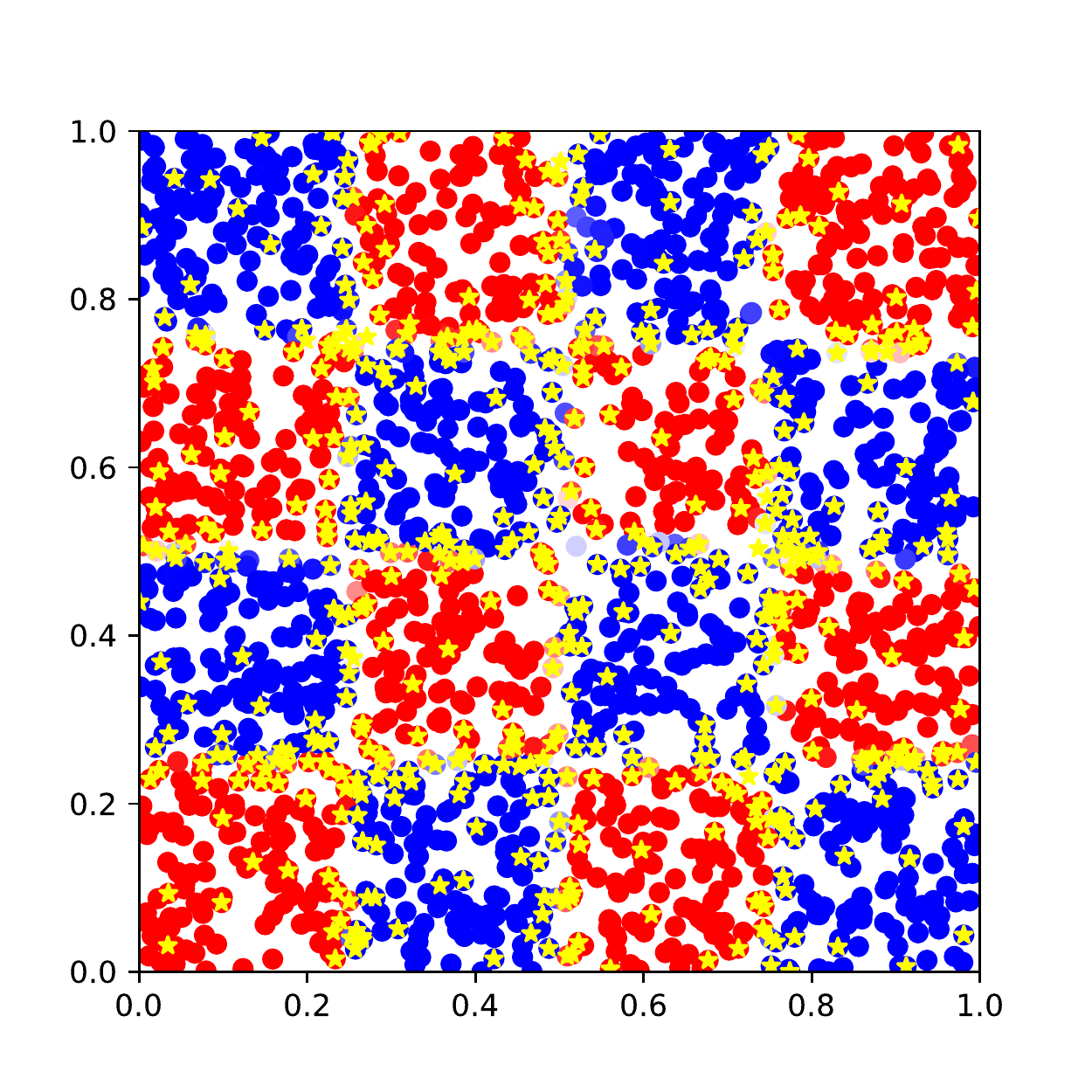}\caption*{Diffusion}
\end{minipage}
\caption{\emph{From left to right}: \emph{(1)} Binary checkerboard pool dataset \emph{(2)} Points queried (in yellow) using \emph{coreset} criteria \citep{sener2017active} \emph{(3)} Points queried using \emph{uncertainty} criterion \citep{gale} \emph{(4)} Points queried using \emph{our} criterion. 
\label{fig:checkerboard_dataset}}
\end{figure}

\begin{figure}[tb]
\center{
\includegraphics[width=.49\textwidth,
keepaspectratio]{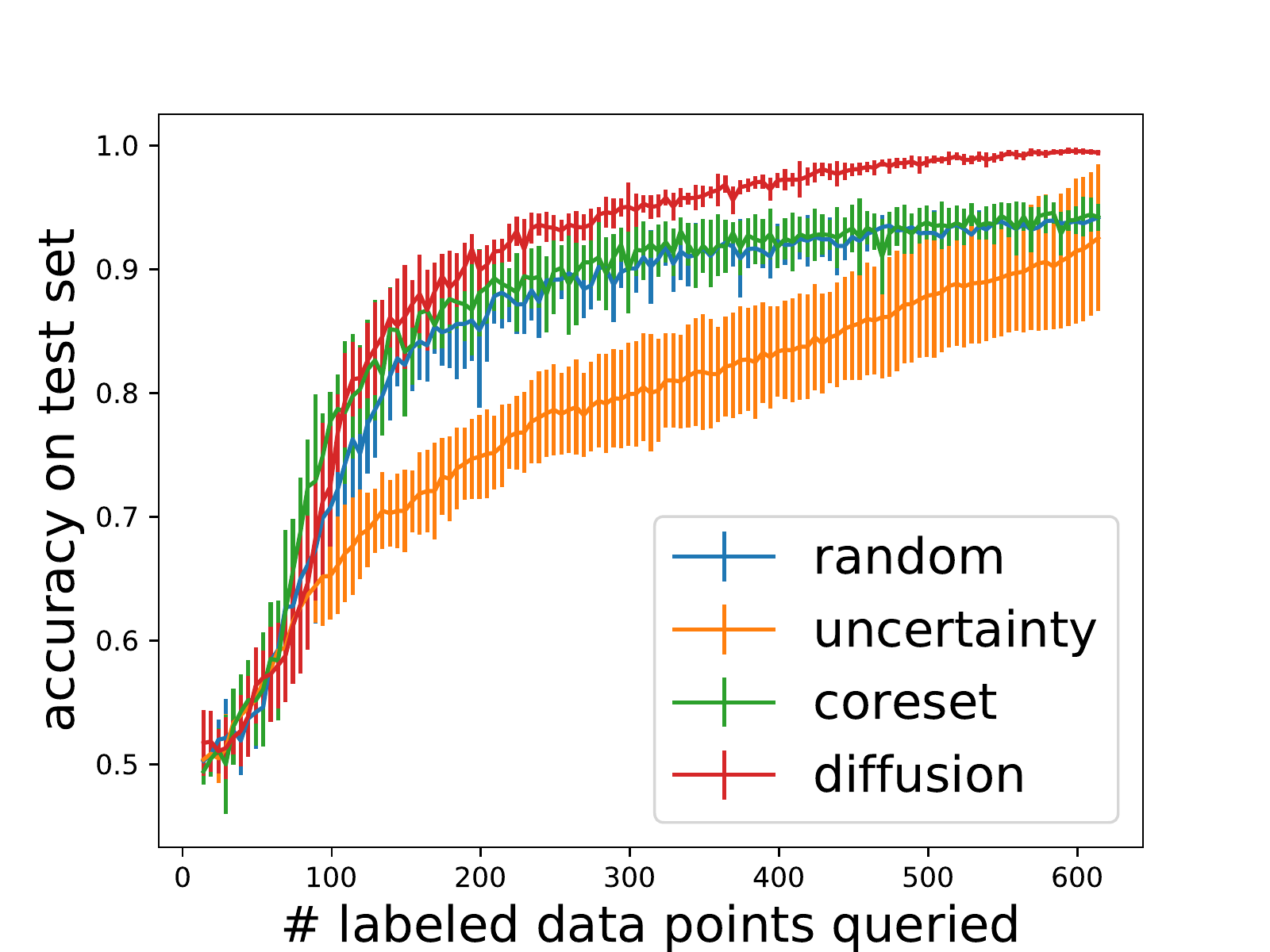}
\includegraphics[width=0.49\textwidth,
keepaspectratio]{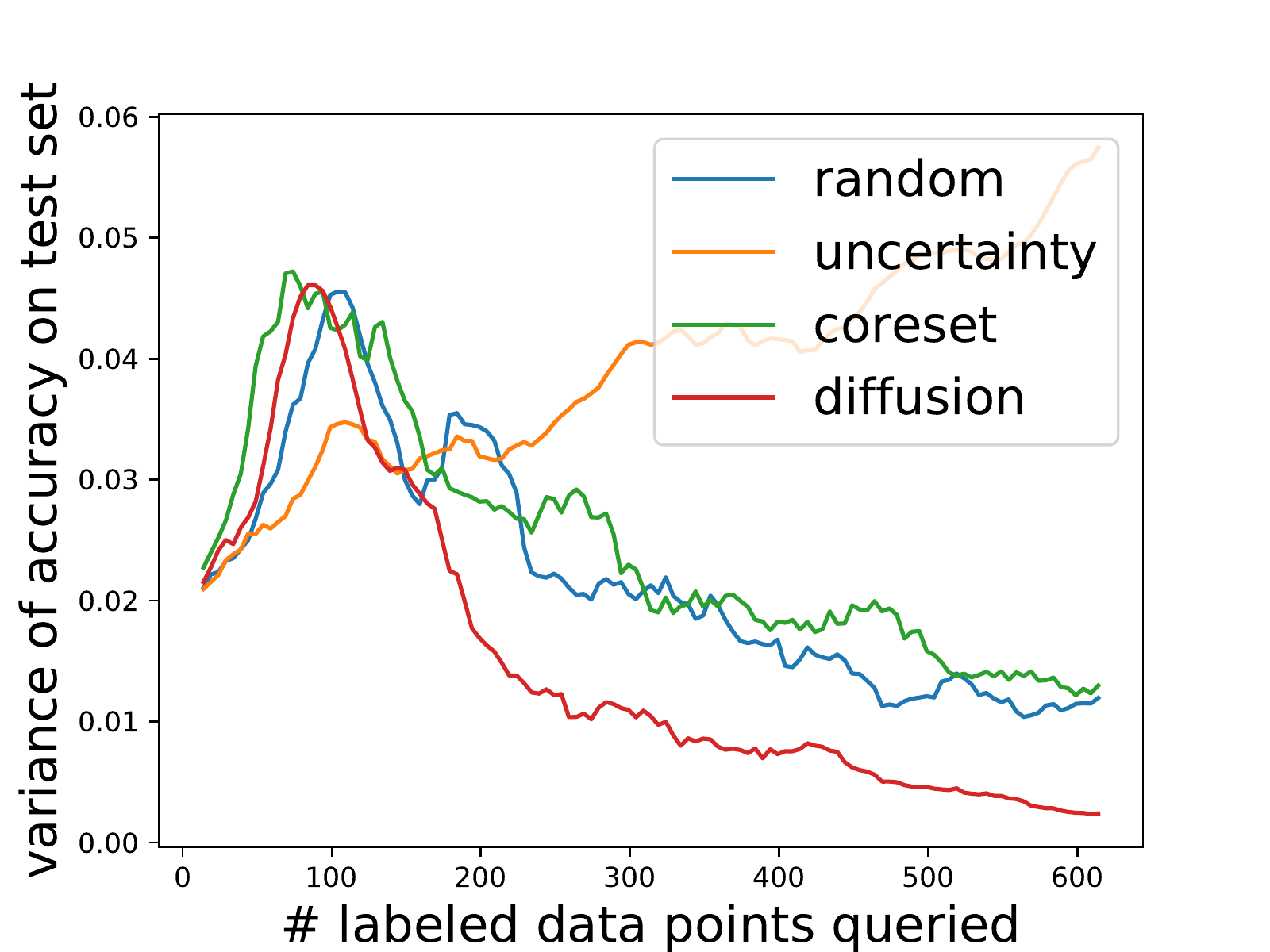}
}
\caption[Caption for LOF]{Performance analysis for active learning on the binary checkerboard set. \emph{Top}: Accuracy versus size of training data set. \emph{Bottom}: Variance of accuracy versus size of training data set.\protect\footnotemark
\label{fig:checkerboard_accuracies}}
\end{figure}

We consider a simple non-separable 2D data set for binary classification. The data set is given as points on a 2-dimensional binary `checkerboard' where each color corresponds to a class (`red', `blue'); see Figure \ref{fig:checkerboard_dataset}. This example demonstrates the utility of refinement vs. exploration criteria in \citep{Baram}.
We consider a pool data set of $\abs{\pool} = 2000$ labeled points drawn from the checkerboard distribution. We start with a training set $\X^0_\ell$ composed of $4$ points randomly drawn from the pool set for each of the two classes (so that $\abs*{\X^0_\ell} = 8$) and train a feed-forward neural network. We run the chosen active learning criterion up to 120 queries with $|B|=5$, and present the first 110 queries in yellow color in Fig. \ref{fig:checkerboard_dataset}. The accuracy and its variance are measured on a separate test data set of $N_\mathrm{test} = 200$ points drawn from the checkerboard distribution in Fig. \ref{fig:checkerboard_accuracies}. All the results are averaged over $5$ runs. Further standard parameter details of the network architecture and training are reported in the supplemental material.

We considered four different selection criteria:
\begin{enumerate}
\item \textbf{Random} - the points are drawn uniformly at random from $X_u$ at each query iteration.
\item \textbf{Uncertainty} - the points where the current model $f_\theta$ is the most uncertain are selected \citep{gale}.
\item \textbf{Coreset} - this is the greedy version of the algorithm introduced in \citep{sener2017active}. It essentially aims to select points in order to form an approximate $\epsilon$-cover of the pool data. 

\item \textbf{Diffusion}
\end{enumerate}

Figure \ref{fig:checkerboard_dataset}-Coreset demonstrates the \emph{exploratory} nature of  \citep{sener2017active}. Coreset aims at uniformly covering the data distribution. Yet, as seen,  the decision boundaries are left unrefined, and therefore its accuracy at the refinement stage is inferior to the diffusion-based criterion.
On the other hand, Figure \ref{fig:checkerboard_dataset}-Uncertainty shows to select points closer to the decision boundary of the current model $f_\theta$. If the model has a discrete understanding of the data distribution, these are close to the decision boundaries of the problem. Nevertheless, at earlier stages of learning, the data queried so far is not enough to cover the distribution and the learned model may be erroneous yet highly confident on unexplored regions. As the model is highly confident on these regions, no points are selected from these regions, leaving them unexplored and causing mis-classification that persists at later stages. At query $110$, we see most uncertainty-based queries concentrate near the detected decision boundary, while boundary segments which have not been explored are completely missed in the classifier.

Finally, Fig. \ref{fig:checkerboard_dataset}-Diffusion illustrates how the \emph{diffusion} criterion operates in two \emph{phases}. In the first, exploratory, phase,
it tends to cover uniformly the distribution and identifies all decision boundaries. At later stages, once the model $f_\theta$ has enough information of the structure of $\D$, it queries labels near the decision boundaries.
Figure \ref{fig:checkerboard_accuracies}-bottom shows that diffusion-based active learning also exhibit smaller variance than all other criteria over different realizations of the same experiments.

\subsection{Benchmarks evaluation}\label{sec:benchmark}
\begin{figure}[tb]
\centering
\begin{minipage}[c]{.49\textwidth}
\includegraphics[width=\textwidth,
keepaspectratio]{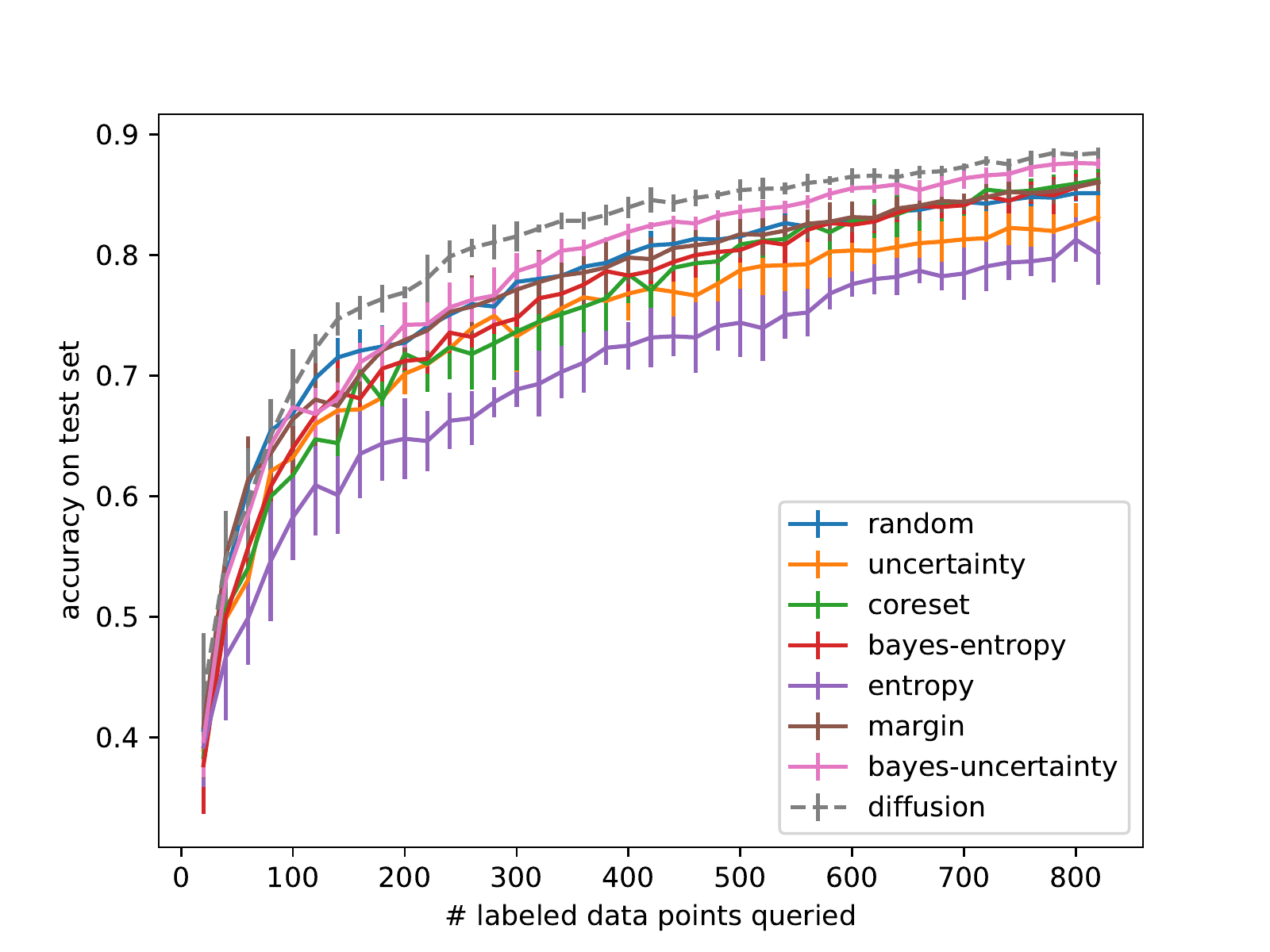}
\end{minipage}
\begin{minipage}[c]{.49\textwidth}
\includegraphics[width=\textwidth,
keepaspectratio]{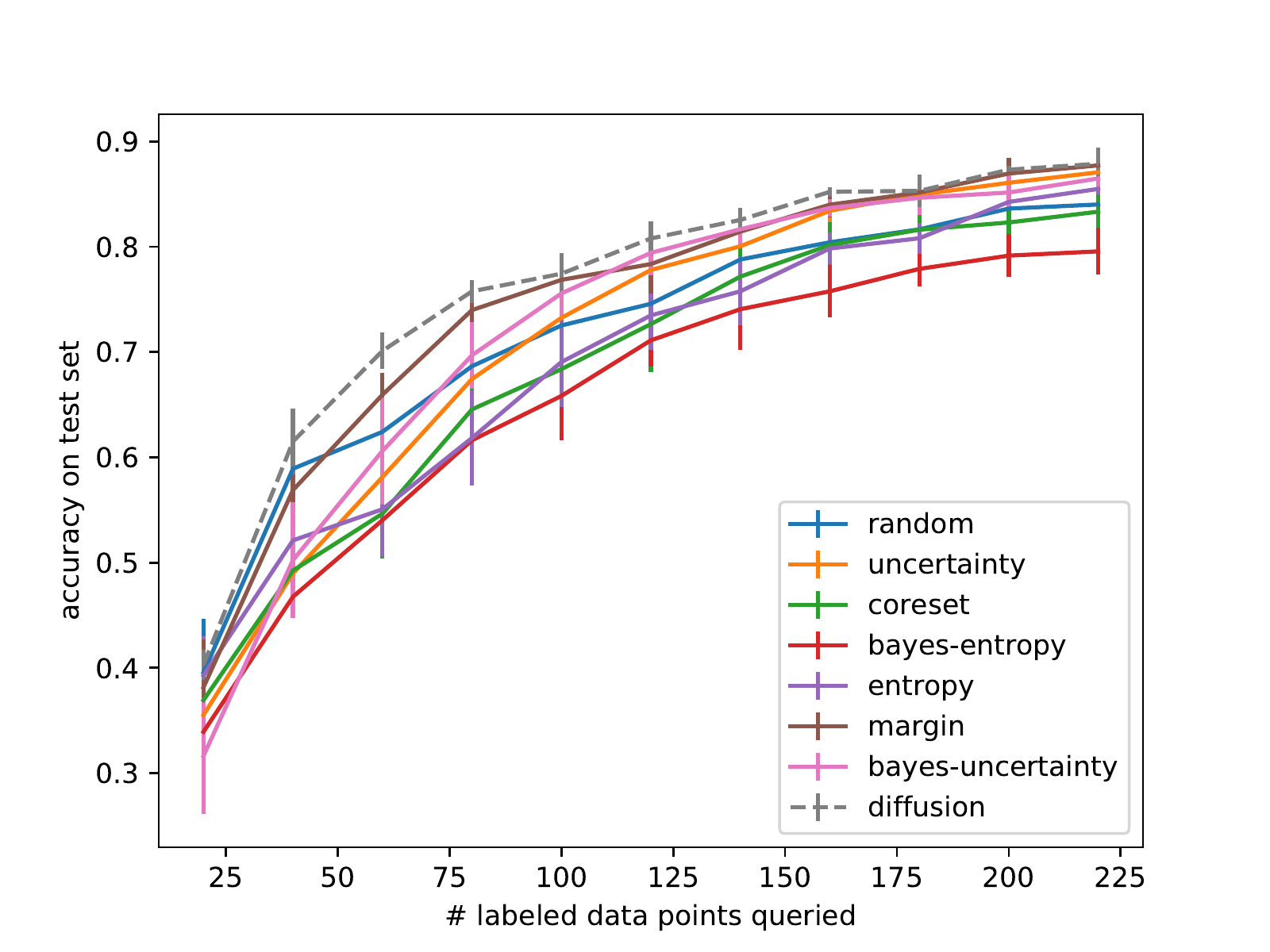}
\end{minipage}
\caption{Left: Results on the MNIST dataset using a fully connected neural net. Right: using a convolutional neural net.  \label{fig:mnist_explore}}
\end{figure}

\begin{figure}[tb]
\centering
\begin{minipage}[c]{.49\textwidth}
\includegraphics[width=\textwidth,
keepaspectratio]{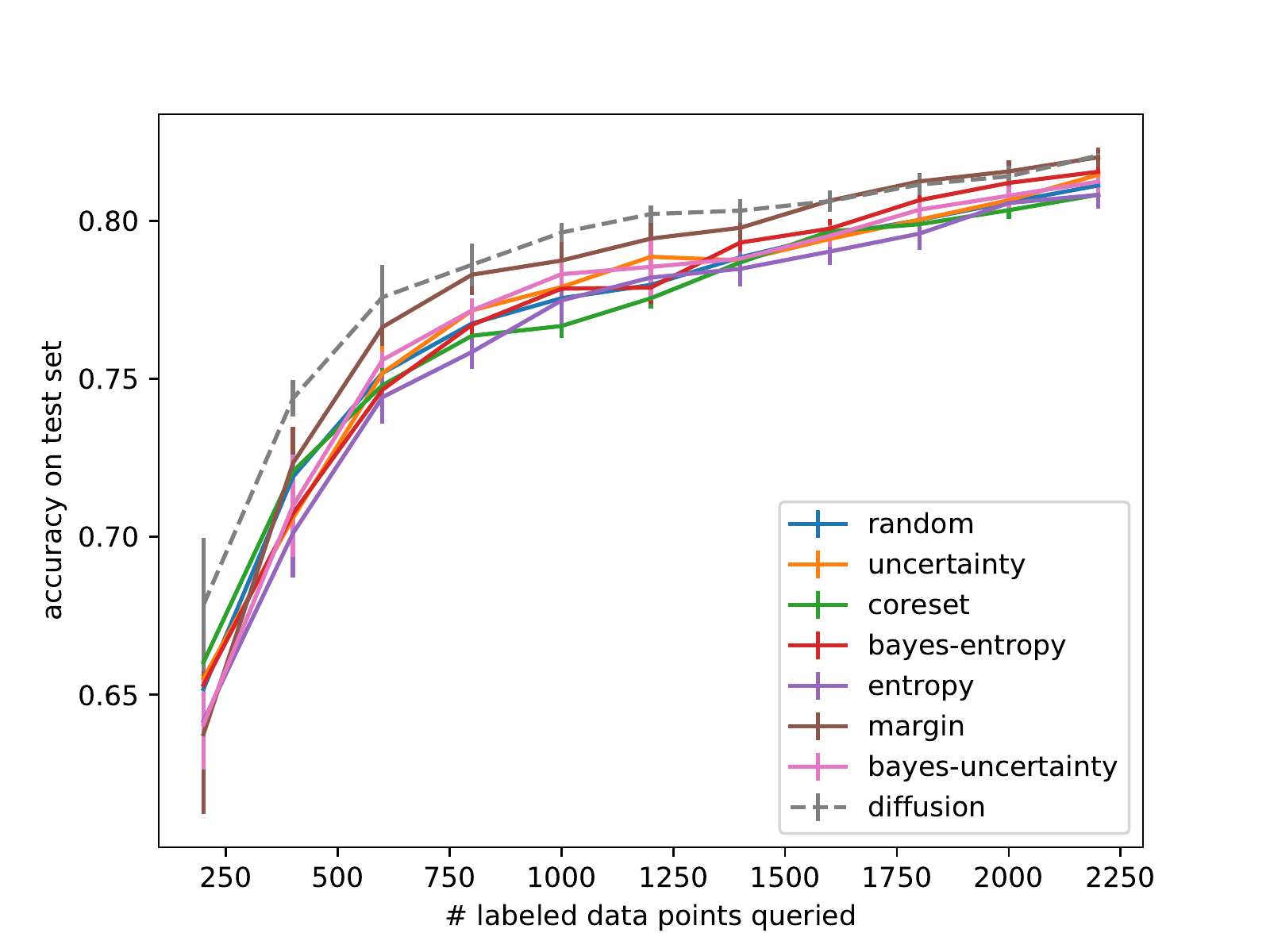}
\end{minipage}
\begin{minipage}[c]{.49\textwidth}
\includegraphics[width=\textwidth,
keepaspectratio]{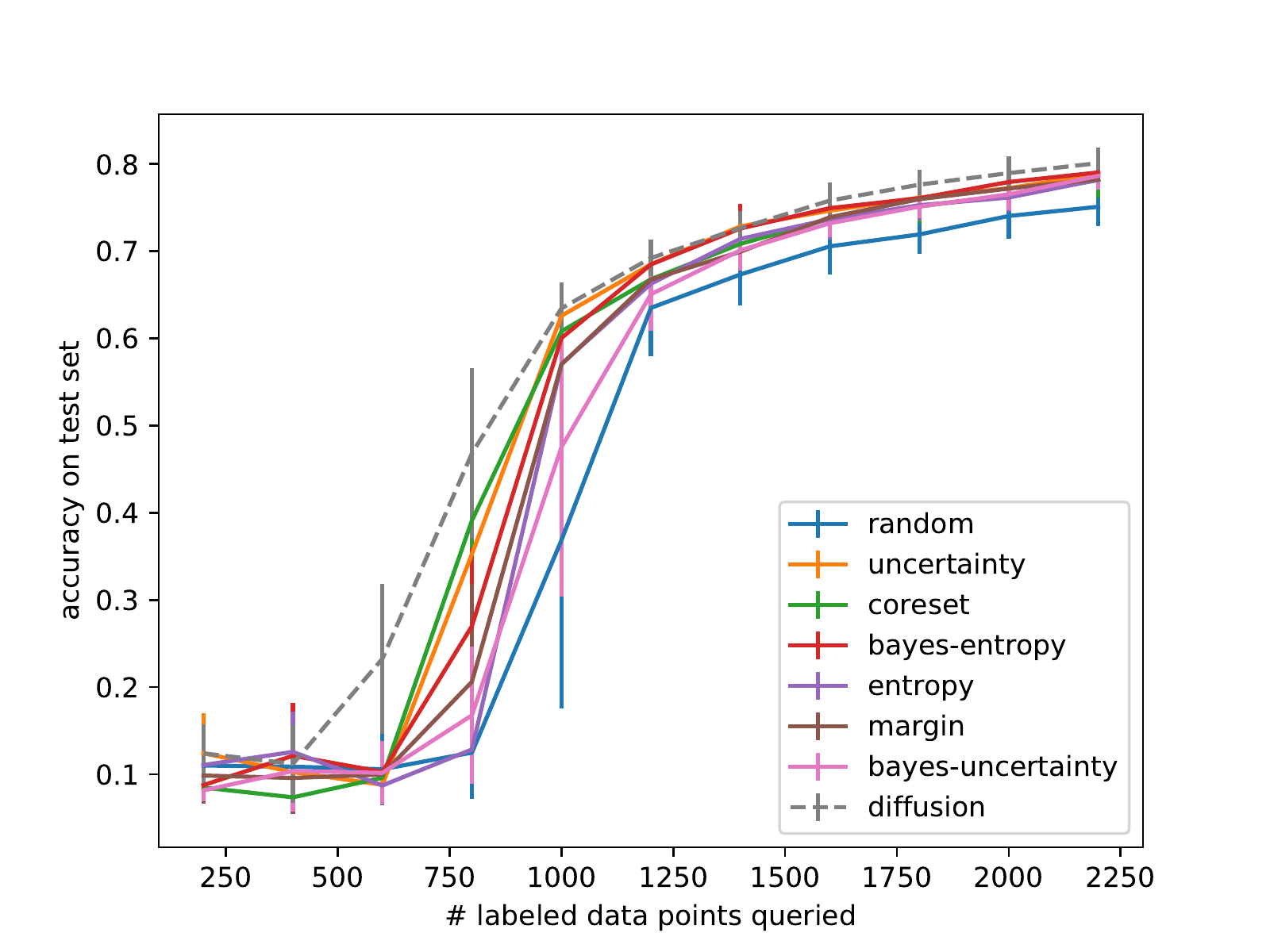}
\end{minipage}
\caption{Left: Results for CIFAR10 data set using a VGG16 network pre-trained on ImageNet. Right: SVHN  data set with VGG16 architecture.   \label{fig:cifar10_svhn_explore}}
\end{figure}

We experiment with the benchmark classification problems: MNIST , CIFAR10, SVHN (figs. \ref{fig:mnist_explore} and \ref{fig:cifar10_svhn_explore}). The advantage of the diffusion-based criterion is especially prominent during the early exploration and the transition to refinement. The competitive accuracy persists into the longer tail of the refinements stage, Where uncertainty criteria is approaching from below. 
\paragraph{Datasets and experimental setups.}

We compared the performance of the diffusion-based active learner with representative methods including \citep{sener2017active} \citep{gal2017deep} and with a range of criteria included in \citep{settles2012active}:

\begin{enumerate}
\item \textbf{Random}
\item Three different \emph{uncertainty}-based criteria:
\begin{enumerate}
\item \textbf{Uncertainty}
\item \textbf{Margin} - the points with lowest output probability difference between the two most probable alternatives are selected (method (2.2) in \citep{settles2012active}). We note that margin is also an explorer-type criterion as it probes representatives points in the early stages of active learning.
\item \textbf{Entropy} - the points with highest entropy in the model’s posterior label distribution are selected (method (2.3) in \citep{settles2012active}).
\end{enumerate}
\item \textbf{Coreset} - \citep{sener2017active}
\item \textbf{Bayesian Dropout} - the algorithm of \citep{gal2017deep}, where dropout is used to obtain an ensemble of models. An uncertainty measure is then evaluated by averaging over the ensemble of models. We considered the following measures of uncertainty:
\begin{enumerate}
\item \textbf{Uncertainty}
\item \textbf{Entropy}
\end{enumerate}
\end{enumerate}

\paragraph{Experimental details.}

For MNIST we consider both fully connected and convolutional models $f_\theta$. For CIFAR10 and SVHN we take $f_\theta$ to be a VGG-16 network \citep{simonyan2014very}. For CIFAR10, the network was pre-trained on ImageNet.
We performed batch queries of size $B = 20$ for MNIST and $B = 200$ for CIFAR10 and SVHN. 
The reported accuracies are averaged over $5$ runs.
All the experiments were performed using PyTorch \citep{paszke2017automatic}. Additional network configuration and training details are reported in Appendix \ref{app:training_details}.


\subsection{Discussion}\label{sec:discussion}
The experiments demonstrate the strong improvement that is obtained by incorporating exploration with refinement through a principled graph-diffusion framework. We observe that during exploration stage `margin' (which may also act as an explorer) as well as `coreset', which are applied in the ambient space, fall behind the diffusion-based learner. We advocate that this is a result of our use of the graph representation which captures the intrinsic geometry of the data points in the labeling space, and accurately models the probability of label assignment in this space. 


Finally, we note the stability of our method vs. all other methods as explicitly shown in Fig. \ref{fig:checkerboard_accuracies}-bottom and in the error bars in Figs. \ref{fig:mnist_explore}-\ref{fig:cifar10_svhn_explore}. This high stability of the diffusion-based method is a key advantage - showing robustness in achieving high accuracy in different initialization (i.e. initially labeled data points) or noisy settings. The  graph representation plays a key role as it denoises the data. On the contrary, we observe that the 'uncertainty` criterion exhibits the highest variance in results as it selects points that do not necessarily represent the distribution (including outliers). Therefore, it is more susceptible to initialization.




\bibliography{main}

\newpage

\appendix

\section{Training details}\label{app:training_details}

In the following we report the hyper-parameter used in the experiments in Section \ref{section:experiments} for each of the data sets.

\paragraph{Checkerboard}

We trained a fully-connected network with $2$ hidden layers of width $30$ each. We optimized using SGD with batch size $1$, learning rate $0.001$ and momentum $0.9$. We ran $100$ epochs after each query. The experiments were run on a pool set of size $2000$. The accuracy was evaluated on a separate test data set sampled from the same distribution ($200$ points).

\paragraph{MNIST}

The fully-connected model used had $2$ hidden layers of width $100$ and $50$ respectively. The convolutional model used was composed by a convolutional layer with 16 channels and a kernel of size $5\times 5$, a MaxPool layer with a kernel of size $2$ and padding $2$, and $2$ hidden fully-connected layers of width $20$ each. For both models, we optimized using Adam \citep{kingma2014adam} with batch size $8$ and learning rate $0.001$. For both models, a BatchNorm \citep{ioffe2015batch} layer was added before each hidden fully-connected layer. We ran $100$ epochs after each query. The experiments were run on a pool set formed by a (balanced) randomly selected subset of the training data set of size $10000$. The accuracy was evaluated on the test data set ($10000$ points).

\paragraph{CIFAR10 }

The network used was a VGG-16 architecture pre-trained on ImageNet. We took the convolutional part of such network and added $2$ fully-connected layers of width $512$ and $20$ respectively. A dropout layer was added after each of these fully-connected layers. Only the fully-connected layers were trained.
We optimized using Adam \citep{kingma2014adam} with batch size $100$. After each query the learning rate was initialized to $0.0003$ and decayed by $0.5$ every $30$ epochs. We ran $100$ epochs after each query. The full training data set ($50000$ points) was used as pool data set for the experiments. The accuracy was evaluated on the test data set ($10000$ points).

\paragraph{SVHN }
The network used was a VGG-16 architecture. 
We optimized using SGD with batch size $50$, learning rate $0.005$, momentum $0.9$ and weight-decay $0.0005$. We ran $50$ epochs after each query. The experiments were run on a pool set formed by a (balanced) randomly selected subset of the training data set of size $20000$. The accuracy was evaluated on the test data set ($26032$ points).

\paragraph{Diffusion}

The following parameters were used for the diffusion algorithm in the experiments presented in Section \ref{section:experiments}. For the experiment with the checkerboard data set (experiment in Figures \ref{fig:checkerboard_dataset} and \ref{fig:checkerboard_accuracies}), we used $T = 4$, $K = 10$ and $P = 1$. For the experiments with the 
MNIST data set (experiment in Figure \ref{fig:mnist_explore}), we used $T = 5$, $K = 10$ and $P = 1$. 
For the experiment with the 
CIFAR10 data set (experiment in Figure \ref{fig:cifar10_svhn_explore}-Top), we used $T = 4$, $K = 20$ and $P = 10$. We also used the technique explained in Section \ref{app:time_change} with $\delta = 0.1$. 
For the experiment with the 
SVHN data set (experiment in Figure \ref{fig:cifar10_svhn_explore}-Bottom), we used $T = 5$, $K = 20$ and $P = 1$. We also used the technique explained in Section \ref{app:time_change} with $\delta = 0.1$ and initialized the signal $\chi^{(0)}$ using the soft-labels information (as explained in Section \ref{app:softlabels}) for each query after the first. 

\paragraph{Bayesian criterion}

In order to perform the active learning queries using the Bayesian criterion, we added a dropout layer after each hidden fully connected layer to the models with no dropout layers.

\section{Variants and enhancements of diffusion-based criterion}\label{app:diffusion_variants}

\subsection{Choosing among points not reached by diffusion \label{app:priority}}

In the first stages of active learning, the diffusion process may not reach all the points. This means that, given a finite time $T$, some of the points $x_i \in \X_u$ may have identically zero propagated vector: $\chi^{(T)}_{i,:} = 0$. Clearly, the query criterion would pick such points (according to \eqref{eq:query}). But in case that there are more of such points than points to query, i.e.
$$
B < \abs*{\bra*{x_i \in \X_u \st \chi^{(T)}_{i,:} = 0}}
$$
we need a criterion to choose $B$ points out of the zero labeled ones. We provide a secondary criterion on top of the diffusion procedure in Algorithm \ref{algorithm:AL}: points are chosen according to an influence criterion
$$
\hatX = \argmax\nolimits^B_{i \in \X_0} \sum_{j \in N(i)} W_{ij}
$$
where we denoted $\X_0 = \bra*{x_i \in \X_u \st \chi^{(T)}_{i,:} = 0}$.

\subsection{Uncertainty signal initialization \label{app:softlabels}}

In Section \ref{section:algorithm}, we described the initialization of the signal $\chi^{(0)} \in \R^{N\times C}$ by using the available labels:
$$
\chi^{(0)}_{i,c} = \begin{cases}
1 & \text{if } i \in \X_\ell \text{ and } c = y_i \\
-1 & \text{if } i \in \X_\ell \text{ and } c \neq y_i \\
0 & \text{if } i \in \X_u
\end{cases} 
$$
Another possibility is to use the softlabels provided by the current model $f_\theta$ to initialize $\chi^{(0)}$, as follows: 
$$
\chi^{(0)}_{i,c} = \begin{cases}
1 & \text{if } i \in \X_\ell \text{ and } c = y_i \\
-1 & \text{if } i \in \X_\ell \text{ and } c \neq y_i \\
2\parr*{f_\theta(x_i)}_c - 1 & \text{if } i \in \X_u
\end{cases} 
$$

\subsection{Dynamically changing diffusion time \label{app:time_change}}

We propose a criterion that reduces $T$ based on the number of points reached by diffusion. This criterion can be incorporated in the enhanced batch query method presented in Section \ref{sec:enhncd_batch}. At each call of \emph{BatchQuery}, the parameter $T$ is initialized to a chosen value $T = T_0$. At iteration $i$ of \emph{BatchQuery}, we count the number $n_0$ of points not reached by the diffusion:
$$
n_0 \doteq \abs*{\bra*{ x_i \in \hatX_u^i \st \chi^{(T)}_{i,:} }}
$$
If $n_0 < \delta N$, we reduce $T$ by $1$: $T = T-1$. Here $\delta \in [0,1]$ is an additional parameter that needs to be fixed.

\section{Proof of Lemma 1}\label{app:proof1}
\begin{proof}
The proof relies on the convergence proof of the Jacobi method \cite{saad}: if $A$ in  \eqref{eq:jacobi} is strictly diagonally dominant then the iteration converges. Since for $L_{uu}$ we have $L_{uu,ij}=\sum_{j \in u} L_{ij}$ , we have that, for all $i \in u$ and $\epsilon>0$, $ L_{uu,ij} +\epsilon I > \sum_{j} \sum_{j \in u} L_{ij}$, i.e. $A \doteq L_{uu}+\epsilon I$ is strictly diagonally dominant. This implies that the sums $s_1,...,s_n$ of the rows of the iteration matrix $B_J$ are smaller than 1. Since $\|A \|_\infty=\max\{s_1,...,s_n\}<1$, we obtain that the spectral radius of $B_J$ is bounded by 1: $\max_i|\lambda_i |\leq \|M \|_\infty<1$. Therefore $B_J$ is a convergent matrix and the Jacobi iteration (\ref{eq:jacobi_iter}) converges.
\end{proof}

\end{document}